%% file: root.tex
\documentclass[letterpaper, 10 pt, conference]{ieeeconf}  

\IEEEoverridecommandlockouts

\usepackage{subfig}
\usepackage{graphicx}
\usepackage{comment}
\usepackage{booktabs}
\usepackage{diagbox}
\usepackage{multirow}
\usepackage{colortbl}
\usepackage{makecell}
\usepackage{amsmath} 
\usepackage[colorinlistoftodos]{todonotes}

\title{\LARGE \bf
An Intelligent Robotic System for Perceptive \\ Pancake Batter Stirring and Precise Pouring
}

\author{Xinyuan Luo$^{1,*}$, Shengmiao Jin$^{1,*}$, Hung-Jui Huang$^{2}$, Wenzhen Yuan$^{1}$
\thanks{This work is funded by Toyota Research Institute (TRI)}
\thanks{*Equal Contribution}
\thanks{$^{1}$Xinyuan Luo, Shengmiao Jin, and Wenzhen Yuan are with the University of Illinois at Urbana-Champaign \{\tt\small xl153, jin45, yuanwz\}@illinois.edu}%
\thanks{$^{2}$Hung-Jui Huang is with Carnegie Mellon University \tt\small hungjuih@andrew.cmu.edu}%
}

\begin{document}

\maketitle
\thispagestyle{empty}
\pagestyle{empty}

\begin{abstract}

Cooking robots have long been desired by the commercial market, while the technical challenge is still significant. A major difficulty comes from the demand of perceiving and handling liquid with different properties. This paper presents a robot system that mixes batter and makes pancakes out of it, where understanding and handling the viscous liquid is an essential component. The system integrates Haptic Sensing and control algorithms to autonomously stir flour and water to achieve the desired batter uniformity, estimate the batter's properties such as the water-flour ratio and liquid level, as well as perform precise manipulations to pour the batter into any specified shape. Experimental results show the system's capability to always produce batter of desired uniformity, estimate water-flour ratio and liquid level precisely, and accurately pour it into complex shapes. This research showcases the potential for robots to assist in kitchens and step towards commercial culinary automation.

\end{abstract}

\input{01Introduction}
\input{02related}

\input{03Perception}
\input{04System}
\input{05Experiment_Results}

\input{06Conclusion}

\addtolength{\textheight}{0cm}   

\vspace{-1mm}
\section*{ACKNOWLEDGMENT}
\vspace{-1mm}

This work was supported by Toyota Research Institute.
The authors would like to thank the entire RoboTouch Lab, especially Amin Mirzaee and Yuchen Mo, for their help with this paper.

\bibliographystyle{unsrt} 
\bibliography{bibliography.bib}

\end{document}

%% file: 01Introduction.tex
\section{Introduction}

The intersection of robotics and culinary arts has opened new avenues for automation in the kitchen, aiming to enhance precision, creativity, and efficiency in food preparation \cite{joe-kitchen} \cite{RoboCook}. However, cooking is still challenging for robots due to the need to manipulate a wide variety of items. Specifically, cooking often involves the use of liquids, which are difficult for robots to handle and perceive because of their changing shapes and complex dynamics. Previous works have explored ways to estimate liquid property inside an enclosed container \cite{discriminate_liquid} \cite{joe-liquid}, but liquid property estimation in open-lid containers poses more challenges. Additionally, although many works aimed at pouring tasks \cite{pourbyfeel} \cite{Pour_Unknown}, culinary applications require additional skills such as controlling the liquid's flow rate and the final shape. These advanced capabilities are crucial for successfully executing tasks in the kitchen.  

\begin{figure}[htbp]
\begin{center}

\includegraphics[scale=0.46]{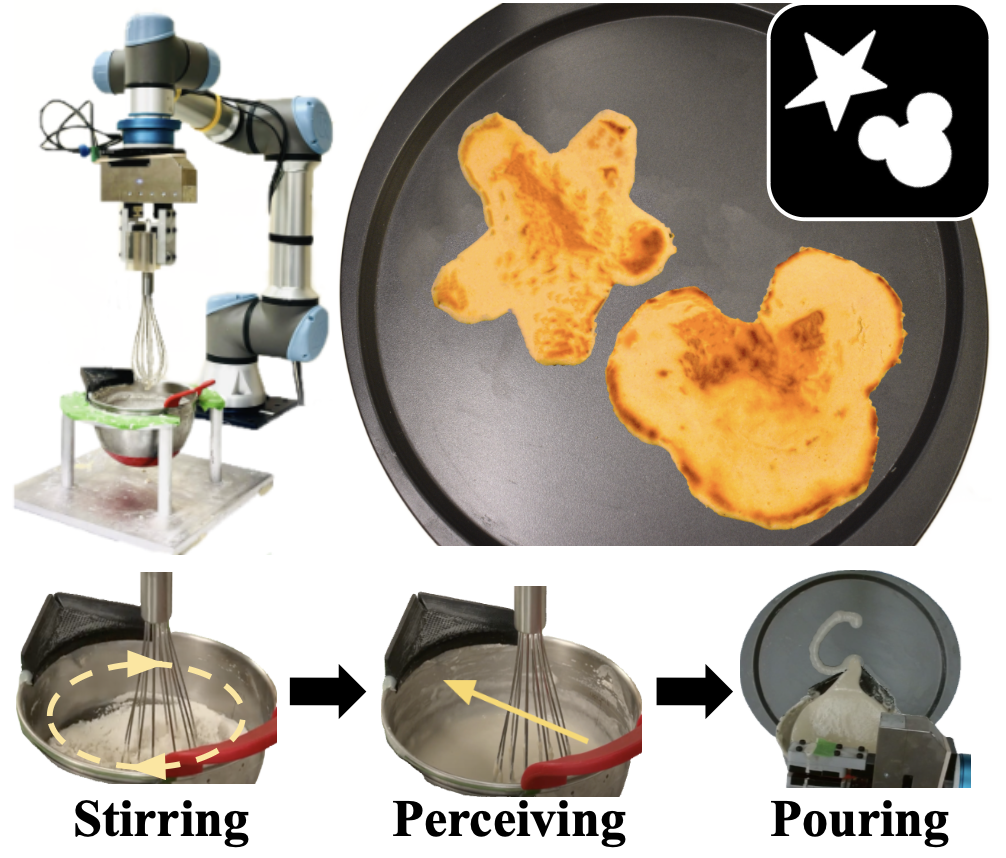}
\end{center}
\caption{We developed an intelligent pancake-making system to perform batter stirring and pouring. Our system can perceive batter and adjust the pouring policy accordingly. }
\label{teaser}
\vspace{-5mm}
\end{figure}
\begin{figure*}[htbp]
\begin{center}
\includegraphics[scale=0.65]
{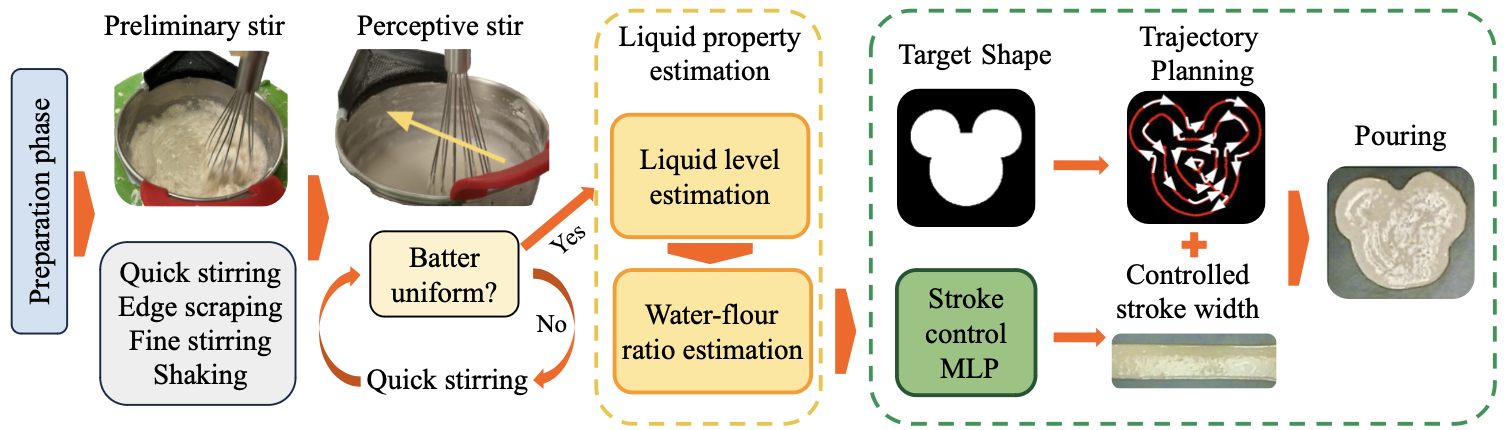}
\end{center}
\caption{Pipeline for our pancake-making system. In the preparation phase, the robot grabs the whisk and then measures the precise position and size of the mixing bowl. Then follows the preliminary stir phase and perceptive stirring until the batter is uniform. Then the robot measures batter's liquid level and water-flour ratio. For pouring, we first plan the trajectory of the robot, then use a stroke control model with liquid level and water-flour ratio as input to regulate the robot's moving speed to maintain the desired stroke width.}
\label{pipeline}
\vspace{-2mm}
\end{figure*}

In this work, our goal is to build an intelligent robotic system that can automate the pancake-making process and make the pancake of any desired shape. Naturally, the process of making pancakes can be divided into two parts: prepare a uniform batter from stirring, and pour it onto a griddle. However, pancake batter presents unique challenges as a liquid due to its variable viscosity — affected by the water-flour ratio—and its non-Newtonian and highly viscous nature. 
Also, to successfully pour pancakes of arbitrary shapes and ensure a uniform distribution of batter, the pouring mechanism must consider the batter's physical properties. This is because batter with varying water-flour ratios exhibits different behaviors, affecting how it should be handled during pouring.
To address these challenges, we incorporate a perception phase into our pipeline along with stirring and pouring, aimed at assessing the properties of the batter to inform and guide the pouring strategy effectively.


We propose a pancake-making system to prepare pancake batter, perceive its properties, and execute precise pouring actions to form pancakes of any desired shape (Fig. \ref{teaser}). We design four different motions in the stirring stage to ensure good quality of batter. Our perceiving stage utilizes a push motion to obtain information on batter uniformity, liquid level, and water-flour ratio. With liquid level and water-flour ratio, we can determine a suitable pouring policy specific to this batter, solving the problems of different batter behaviors. 
We then implement a Multi-Layer Perceptron (MLP)  model to estimate the robot arm movement speed from water-flour ratios to achieve consistent line strokes of the desired width. With a trajectory decomposition algorithm, we can turn a binary figure input into a usable trajectory for the robot to pour the batter input into the input shape. This integrated approach ensures that our system can manipulate different batters with consistency, paving the way for automated pancake creation with high precision and adaptability. 

Our system's key contribution is its ability to evaluate the liquid level, and water-flour ratios of pancake batter after making sure the batter is uniformly mixed, coupled with an adaptive strategy for precisely pouring various batters. 
Experimental results show that our method can achieve uniform batter for all 15 trials with different bowls. Our method can also estimate the water-flour ratio within $9.6\%$ error and liquid level within $3.88\%$ relative error. Our resulting line stroke width has only a relative error of $9.5\%$ and a variance of $9.6\%$ across several different water-flour ratios. Finally, we demonstrate that we can make pancakes of any desired shape. Our proposed system enhances both the functional and creative aspects of cooking robotics.

%% file: 02related.tex
\section{Related works}
\subsection{Cooking Robots}

The potential of robots in culinary tasks has been explored in some previous works. Kumagai et al \cite{assist_cooking} developed a system that can search online for receipts and suggest the next move using voice and gesture in a kitchen setting. However, building a robotics system that can be generalized for all cooking tasks is challenging. The above work does not use robots to perform actual cooking. Many works targeted only specific tasks in culinary scenarios. Liu et al. \cite{stirfry} developed a bi-manual robot system to perform the Chinese cooking style of stir-fry. Shi et al. \cite{RoboCook} taught robots to use tools to perform long-horizon tasks such as dumpling making. None of those systems possessed the ability to make pancakes. In this work, we specifically explore pancake-making robots.  A more relevant work by Beetz et al. \cite{pancake} directly addressed the pancake-making potential of humanoid robotics without any help from humans around. Their work focused on flipping the pancake and robot cooperation with passing batter from a fridge. Our work presents an entire pipeline that includes stirring the batter and pouring it to make various desired shapes.

\subsection{Liquid Property Estimation}

Understanding the physical properties of liquids is essential for 
manipulating them in an accurate way.
The composition and formula of pancake batter not only affect the taste and texture of the pancakes \cite{pancake2} but also play a significant role in determining the pouring policy. Many works have studied liquid properties estimation in both simulated environments \cite{li2018learning} \cite{xian2023fluidlab} and real-world \cite{liquid_precption22}. Matl et al. \cite{matl} used haptic signals during bottle rotation and a physics-based approach to predict the liquid viscosity. Huang et al. \cite{joe-liquid} studied the tactile data with an impulse action on an enclosed bottle and showed how that could be used to estimate liquid properties.
In Kitchen Artists \cite{joe-kitchen}, Huang et al. showed how the perception of liquid properties can guide the squeezing policy of different sauces, focusing on more viscous liquids. 
Our research shares similarities with Kitchen Artists \cite{joe-kitchen} in measuring and manipulating viscous liquids; however, our focus extends to liquids with non-Newtonian properties. Additionally, unlike most studies that involve enclosed containers, our work requires interacting with an open mixing bowl, where the interaction space for perception is more limited. 
A more related study, the Stir-to-Pour project \cite{stir-to-pour} explored liquid dynamics modeling through stirring and visual feedback to fit a simulated liquid model. Our approach also involves an interactive motion for perceiving liquid properties but relies on force torque sensors without simulation to inform our pouring policy.
\subsection{Liquid Pouring}

Robot pouring is the most common task involving the manipulation of liquids, and in most cases, researchers focus on controlling the amount of low-viscosity liquid that is dispensed into a container
\cite{pourbyfeel} \cite{Pour_Unknown} \cite{do2018accurate}. Schenck and Fox \cite{pour17} utilized RGB images to estimate liquid volume in containers for feedback control in pouring. Liang et al. \cite{Liang_2020} used audio and haptic feedback to control the precise amount of liquid poured into a container. Babaians et al. \cite{pournet} introduced an RL-based method to obtain a more precise result for the desired amount in a certain container. Matl et al. \cite{matl} explored precise pouring in an open-loop manner with information obtained by perceiving the properties of the liquid. 
Compared to those works, our approach focuses on dispensing high-viscosity liquid with controllable flow rates. 
Huang et al.'s Kitchen Artist \cite{joe-kitchen} has a similar goal to ours to control the flow rate of viscous liquid but with a squeeze bottle instead of pouring. Our setup poses a more challenging setting for liquid manipulation.


%% file: 03Perception.tex
\section{Perceiving Liquid Batter \\and Precise Pouring}
\label{perceive liquid batter and pouring}

In this section, we discuss the methodologies for assessing liquid properties and establishing a precise pouring strategy, with the ultimate goal of crafting pancakes in any shape the user desires, as shown in Fig. \ref{pipeline}. The inherent high-viscosity and non-Newtonian properties of pancake batter introduce complexities in both its measurement and manipulation. To address this, we employ a push motion that exerts a constant force on the batter to gather data on its uniformity, liquid level, and water-flour ratio. This approach effectively neutralizes the impact of the batter's non-Newtonian characteristics with constant force, ensuring the measurements are comparable. Our approach incorporates the estimation of these vital properties to devise a pouring policy that controls the stroke width with high precision.
\subsection{Liquid Property Estimation}
\label{Liquid property estimation}
We aim to estimate the mixture's uniformity, liquid level, and water-flour ratio of batter. Uniformity assessment ensures the batter's quality, confirming its readiness for pouring. The measurement of the batter liquid level is utilized for determining the water-flour ratio and also the initial angle when pouring. 
The water-flour ratio would be used to determine our pouring policy. 
In our approach, the robot horizontally pushes the whisk in the batter, measures the resistance torque of the y-axis using a wrist-mounted Force-Torque (F/T) sensor, and uses it to infer the liquid properties. Our insight is that the varying viscosities and depths of the batter result in distinct resistance torques during the horizontal pushing process.
We set the pushing speed to $2.5$ cm/s and the pushing distance to $5$ cm. 


\subsubsection{Uniformity}
\label{Uniformity}
The uniformity of liquid batter refers to the evenness of its composition throughout, with respect to ingredients being thoroughly mixed and resolved.
The goal of the uniformity check is to control the robot to stop stirring once the batter achieves uniformity. 
To achieve this, we measure the average torque of a push motion after $50$ rounds of quick stirring in the perceptive stiring stage, which is defined as a trial. 

\begin{figure}[thbp]
\begin{center}
\includegraphics[scale=0.45]{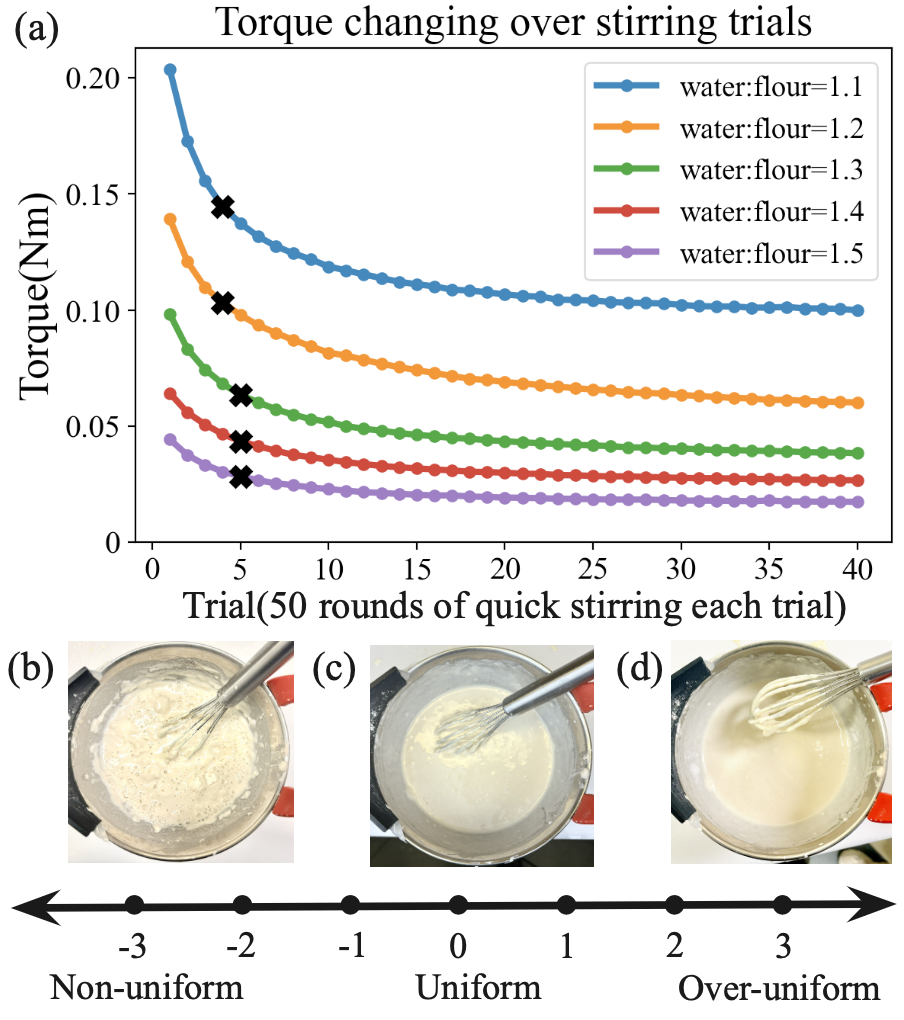}
\vspace{-7pt}
\end{center}
\caption{Batter uniformity analysis. (a) Reduction in resistance torque due to increasing uniformity during the stirring of five batches of batter, with water-flour ratios ranging from $1.1$ to $1.5$. Black crosses mark the points of uniformity as assessed by human observers, who use a scale ranging from $-3$ to $3$: (b) non-uniform example($-3$), (c) uniform example($0$), and (d) over-uniform example($3$).}
\label{torque vs vision}
\end{figure}

Fig. \ref{torque vs vision}(a) shows the changes of torque throughout the stirring process when stirring five batches of batter with a water-flour ratio ranging from $1.1$ to $1.5$. 
The resistance torque continues to decrease over the $30$ minute stirring period, indicating that the batter viscosity keeps decreasing and never converges. This phenomenon could be attributed to the batter becoming finer as a result of constant stirring. However, humans typically assess batter uniformity after only $4$ to $5$ trials of quick stirring (approximately $160$ to $200$ seconds) as shown in Fig. \ref{torque vs vision}(c). To keep our system efficient, we want to prevent stirring to the over-uniform state as shown in Fig. \ref{torque vs vision}(d).

To control the robot to stop stirring at the appropriate time, we set a threshold for the changing rate of the measured torque based on human labeling. This label is obtained by the human observer's rating on each trial's resulting batter, with the scale shown in Fig. \ref{torque vs vision}.
We define the point where human observers' rating average is $0$ to be our uniformity threshold, represented by the black crosses in Fig \ref{torque vs vision}(a). We aim for the robot to stop stirring at a point that aligns closely with these identified moments. This is achieved by comparing the difference in torque between consecutive trials against this threshold, thereby determining the uniformity of the batter.

\vspace{-5pt}
\subsubsection{Liquid Level}
\label{sec:depth}

\begin{figure}[t]
\begin{center}
\vspace{2mm}
\includegraphics[scale=0.4]{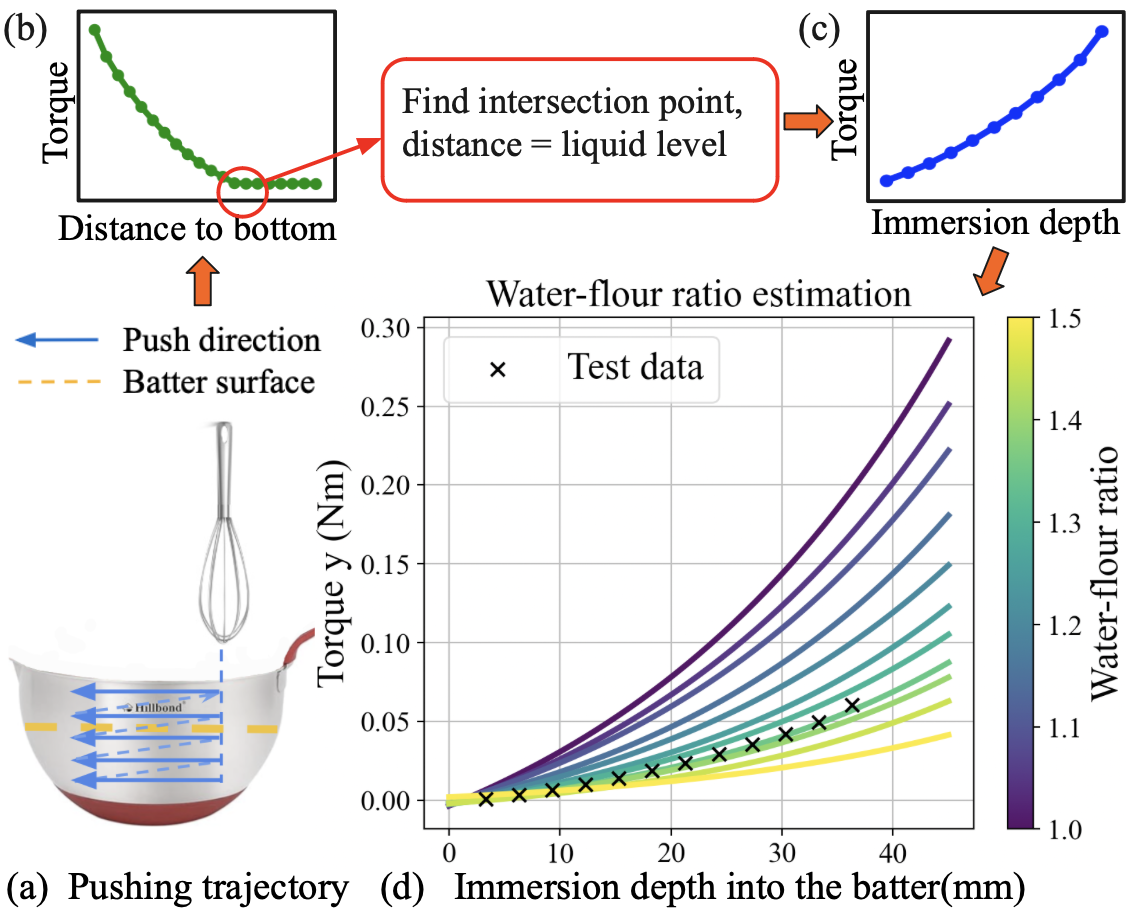}
\end{center}
\caption{This figure shows the pipeline of liquid level and water-flour ratio estimation. (a) The trajectory of the pushing motion of each trial is represented by blue lines. Torque measurements are acquired from each pushing motion, resulting in a Torque-Distance to bottom curve (b). This curve is used to estimate liquid level. Following this, a Torque-Immersion curve (c) is derived from the estimated liquid level. Finally, the water-flour ratio is estimated using this data through our data-driven model (d).}

\label{liquid property estimation overwiew}
\end{figure}

The liquid level is defined as the distance from the bottom of the bowl to the surface of the batter. 
We employ multiple push motions in an increasing height sequence.
Fig. \ref{liquid property estimation overwiew}(a) illustrates the trajectory of this motion. After each pushing motion, we obtain an average torque reading. We show the relation between torque and the distance from the tip of the whisk to the bottom of the bowl, illustrated in Fig. \ref{liquid property estimation overwiew}(b).
The slope portion corresponds to pushing in the batter while the flat portion corresponds to pushing in the air where the torque is around $0$ N.

To differentiate between a push occurring in the batter versus in the air, we set a threshold for the variation in torque measurements between successive pushes. By fitting the torque data for both batter and air pushes to linear functions, we then identify the intersection of these fitted lines as the precise point where air transitions to batter. This intersection point shows the liquid level of the batter.



\subsubsection{Water-flour Ratio} 
Under the assumption of a uniformly mixed batter, a relationship exists between the water-flour ratio and the viscosity of the mixture. Liquids with higher viscosities offer more resistance during stirring, allowing us to develop a data-driven model to estimate the water-flour ratio based on the observed resistance torque. 
We collect torque data by pushing into various batches of liquid batter with different water-flour ratios and immersive depths to measure the water-flour ratio.
We found that greater immersive depths and smaller water-flour ratios yield larger torques, which is consistent with expectations.

%
To estimate the water-flour ratio of an unknown batter, our model takes the ``Torque vs. Immersion depth" curve 
obtained in section \ref{sec:depth} as input. Then it finds the closest two fitted curves that minimize mean square error (MSE) and estimates the water-flour ratio by calculating the weighted average value as $r_E=(M_1r_1 + M_2r_2)/(M_1 + M_2)$. Where $r_E$ is the estimated water-flour ratio, $r_i$ is the water-flour ratio label. $M_i$ denotes the MSE between predicted and measured torque, computed as $M_i = \frac{1}{n} \sum_{j}(\hat{T_j} - T_j)^2$, with $n$ representing the total pushing trials count, $j$ indexes each data point in the ``Torque-Immersion depth'' curve and $\hat{T_j}$ denoting predicted torque in the dataset with $T_j$ being measured torque.



\subsection{Precise Controlling of Pouring}
\label{stroke}

We want to control the width of the lines which is influenced by two key factors: the batter flow rate and the robot arm moving speed. 
While using visual feedback for intuitive control seems logical, real-time control based on visual feedback is challenging.
We, therefore, propose an open-loop control using the water-flour ratio and depth obtained from perception to predict the batter flow rate. We develop a model to incorporate estimations from the perception stage to refine our pouring policy.
Overall, our approach is divided into two main stages: controlling the batter flow rate and establishing a model to predict the robot's movement speed from a desired batter stroke width. 

\subsubsection{Precise Control of Flow Rate }
To achieve a precisely controlled batter-pouring process, we aim to control the robot's pouring speed with consistency and accuracy. In our method, the robot first grabs the bowl handle, and we control the rotation of the wrist joint to pour out the batter. Through experimentation, we discover that maintaining an angular speed of $0.007$ radian per second enables us to achieve a uniform flow rate across various water-flour ratios. By sticking to this speed, we can control the flow rate in a quasi-static manner.
\subsubsection{Speed and Time Control}
With a constant flow rate of the batter poured out, we then control the motion speed of the robot to form an even batter stroke with well-controlled widths, which is the basic requirement to form an arbitrarily shaped pancake.
We train an MLP model to predict the robot's moving speed from the water-flour ratio of the batter and the desired stroke width. 

In another more traditional case, we aimed for the robot to create a round pancake by pouring the batter from a stationary position. Here, we supervise the pouring duration to determine the pancake's size. 
We train another MLP model to predict this pouring time. 

%% file: 04System.tex
\section{Pancake Robot Pipeline}
The overall pipeline of pancake making is divided into four sub-tasks: preprocessing, stirring, perception, and pouring. We have discussed our approaches to perception and controlling poured line widths in Section \ref{perceive liquid batter and pouring}, and in this section, we will introduce other technical components of our pancake making system. 
\vspace{-4pt}
\subsection{Preparation Process}
This process includes taking the whisk and measuring the precise location and the size of the mixing bowl. The whisk is initially located on a fixed whisk shelf (Fig \ref{setup}). The robot automatically picks up the whisk, making sure to grasp it at the same spot each time. This consistency ensures that the force and torque readings we gather are comparable and reliable.
The robot first holds the whisk to go forward, backward, left, and right to explore the bowl edge. The force/torque (F/T) sensor detects a force exceeding 5 N on the x-y plane during these exploratory movements, the robot halts and logs the displacement. With those results, the center and radius of the circle can be inferred and will be used as the precise location and size of the bowl. 


\subsection{Stirring}
\label{sec:stir}
We developed four motions to mix water and flour: \textit{quick stirring, fine stirring, edge scraping,} and  \textit{whisk shaking.} Table \ref{stirring} shows parameters in detail.

\begin{table}[ht]

\centering
\begin{tabular}{|l c c c c|}
\hline
 & \textbf{Speed} & \textbf{Depth} &\textbf{Radius}  & \textbf{Rotation}\\
\hline
\hline
Quick stirring  &  $15.7$ rad/s & $h-2$ mm & $r-2$ mm & No\\
\hline
Fine stirring & $6.28$ rad/s  &  $h-5$ mm & $r$ mm & Yes\\
\hline
Edge scraping & $3.14$ rad/s  &  $h-15$ mm & $r+2$ mm & Yes\\
\hline
Shaking & $8$ Hz  &  $h-5$ mm & N/A & No \\
\hline
\end{tabular}

\caption{Parameters of the four motions, where $h$ denotes the liquid level and $r$ denotes the measured radius of the bowl. Rotation is a supplementary movement enabling the whisk to spin back and forth along its axis, akin to the motion of an egg beater.}
\label{stirring}
\vspace{-2mm}
\end{table}

\subsubsection{Quick Stirring} This motion is designed for mixing water and flour at high speed to achieve uniformity. The motion trajectory's radius is smaller than the bowl, thus it doesn't make contact with it. 

\subsubsection{Fine Stirring} This motion is designed to handle the area that quick stirring motion is unable to reach. At the initial stage of stirring, flour easily accumulates in those areas. If not dispersed early in the accumulation, it will form hard-to-dissolve flour clumps. 


\subsubsection{Edge Scraping} This motion is designed to scratch the batter stick to the border of the bowl, which typically happens at the initial stage of stirring. 
The whisk will contact with the bowl's edge and have small deformations. 
The trajectory's height is $15$ mm above the bottom of the bowl to scrape the flour above the liquid surface.

\subsubsection{Whisk Shaking} This motion is designed to get rid of unresolved batter stuck in the space within the whisk. 
We command the whisk to translate back and forth within the batter at a frequency of $8$ Hz and at the distance of $5$ cm.

Our stirring process includes a preliminary stirring phase to mix the raw material into a batter mixture and a perceptive stirring phase to achieve higher batter uniformity. In the preliminary stirring phase, we execute the following series of actions over a 90-second period: quick stirring, edge scraping, fine stirring, and whisk shaking. In the perceptive stirring phase, we execute $50$ rounds of quick stirring and assess the mixture's uniformity by conducting the push test repeatedly until the batter reaches uniform. Following this, we proceed to the liquid property estimation sub-task (Section \ref{Liquid property estimation}).

\vspace{-4pt}
\subsection{Pouring Control and Planning}

This section outlines our approach for precisely controlling the pouring of batter and planning varied trajectories for different input shapes, aiming to create pancakes of arbitrary shapes. Our method includes an initial angle control module to find an optimal initial angle and a trajectory planning module that shapes the poured batter into the desired forms. 


\subsubsection{Control of Initial Angle}

\begin{figure}[htbp]
\begin{center}
\includegraphics[scale=0.35]{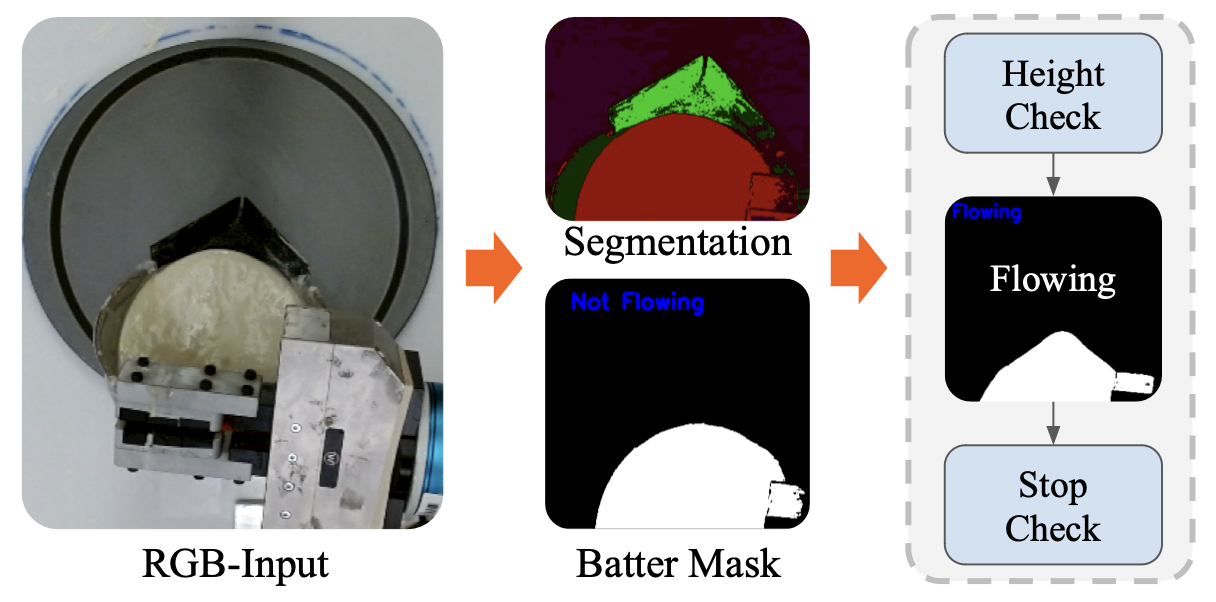}
\vspace{-7pt}
\end{center}
\caption{Viusal Feedback Control Pipeline. When the vertical distance between the top and bottom pixels of the batter segment starts to increase, it means the batter begins to flow onto the spout. When the distance stops to increase, it signals the batter has flowed to the end of the spout.}
\label{vcl}
\end{figure}

Another source that influences the flow rate of the batter is the initial angle. In Section \ref{stroke}, we discussed how to maintain a flow rate, but the initial flow rate can vary if we change the initial angle. The goal of this module is to find an optimal initial angle so that batter with the same water-flour ratio and different liquid levels can have a constant initial flow rate.
We design the spout in Fig. \ref{setup} to help regulate batter flow. We first use the liquid level to estimate a starting angle smaller than what it should be. Our system utilizes visual input to monitor the flow of the batter, with a KMeans segmentation to isolate the batter. When the batter is detected to drip from the spout, we initiate the pouring movement, as shown in Fig. \ref{vcl}.


\subsubsection{Trajectory Planning}

To form an arbitrarily shaped pancake, we create an algorithm that can decompose a shape from a binary image into a batter-pouring trajectory for the robot. Fig. \ref{pipeline} shows an example of the trajectory we generate. Our algorithm contains two modes for generating trajectories of different shapes: 
\begin{itemize}
  \item For an enclosed shape, our algorithm uses the shape's edge as the initial loop and erodes the shape to generate trajectories for inner loops.
  \item For non-enclosed lines, our algorithm skeletonizes the image to form the trajectory 
  and then refine the shape using the Minimum Spanning Tree algorithm.
\end{itemize}

%% file: 05Experiment_Results.tex
\section{Experiments and Results}
In this section, we test the perception and pouring component of our pancake making system. For our perception system, we evaluate the precision in estimating uniformity, water-flour ratio, and liquid level of the pancake batter in circular containers with various sizes. For our pouring system, we test our method on the task of controlling the diameter of the round pancake and the width of drawn line strokes. 


\begin{figure}[htbp]
\begin{center}
\vspace{2mm}
\includegraphics[scale=0.62]{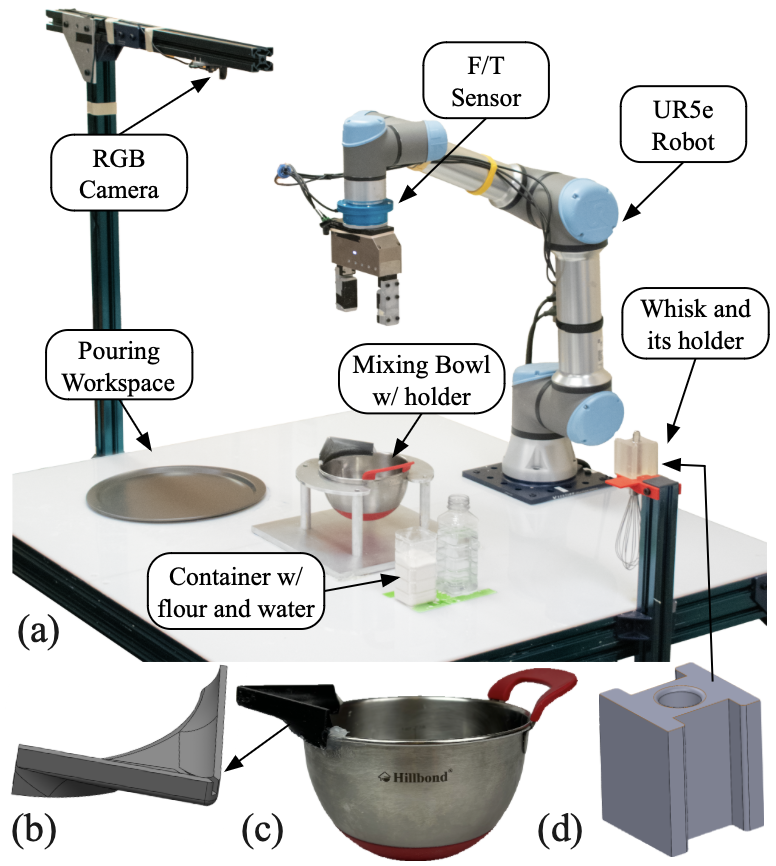}
\end{center}
\caption{(a) System setup overview of our pancake-making robot. (b) Specialized pouring spout. (c) Mixing bowl equipped with the specialized pouring spout. (d) Specialized whisk gripping holder.}
\label{setup}
\vspace{-5mm}
\end{figure}

\subsection{Experiment Setup}
Fig. \ref{setup} shows our experimental setup. Our system uses a 6-DoF UR5e Robot arm by Universal Robotics. This arm is equipped with a WSG-50 2-fingered gripper from Weiss Robotics with aluminum fingers. Attached to the robot's wrist is a 6-axis NRS-6050-D80 F/T sensor from Nordbo Robotics with a sampling rate of 1000 Hz. To ensure stability during mixing, we mount the mixing bowl on a holder and place a shelf equipped with a 1920x1080 resolution RGB camera above it. For the stirring experiments, we utilized a default small bowl with an 8.3 cm radius and 1100 ml volume, as well as a large bowl with a 10.5 cm radius and 2200 ml volume. 



\subsection{Liquid Property Estimation}
In this section, we assess our liquid property estimation approach, which includes uniformity verification, liquid level estimation, and water-flour ratio estimation. 
\subsubsection{Data Collection}
We collect training data for building the water-flour ratio estimation model and collect test data for uniformity check, liquid level, and water-flour ratio estimation. In both sets, stirring is performed by the robot. For the training set, we only collect data in the small bowl. The test set employs both the small bowl and the large bowl, demonstrating our method's generalizability.



For the training set, we collect data from batters with a water-flour ratio varying from $1.0$ to $1.5$ in increments of $0.05$. The water quantity used for each bowl of batter is controlled around $300$ ml, along with the corresponding amount of flour by weight. The robot is programmed to execute $5$ cm push motions from a height of $3$ mm to $63$ mm above the bottom in $3$ mm increments, following the trajectory depicted in Fig \ref{liquid property estimation overwiew}(a), while torque data is recorded.
In total, $11$ trials of batter are made and $660$ data points are collected to build our model.

For the test set, $10$ and $5$ trials of batter are prepared in small and large bowls, respectively, with water-flour ratios randomly selected between $1.0$ and $1.5$ and total weights varying from $450$ to $700$ g. For each batter trial, uniformity is assessed by human observers, followed by evaluation of the accuracy of liquid level and water-flour ratio estimations.


\begin{figure}[htbp]
\begin{center}
\vspace{2mm}
\includegraphics[scale=0.38]{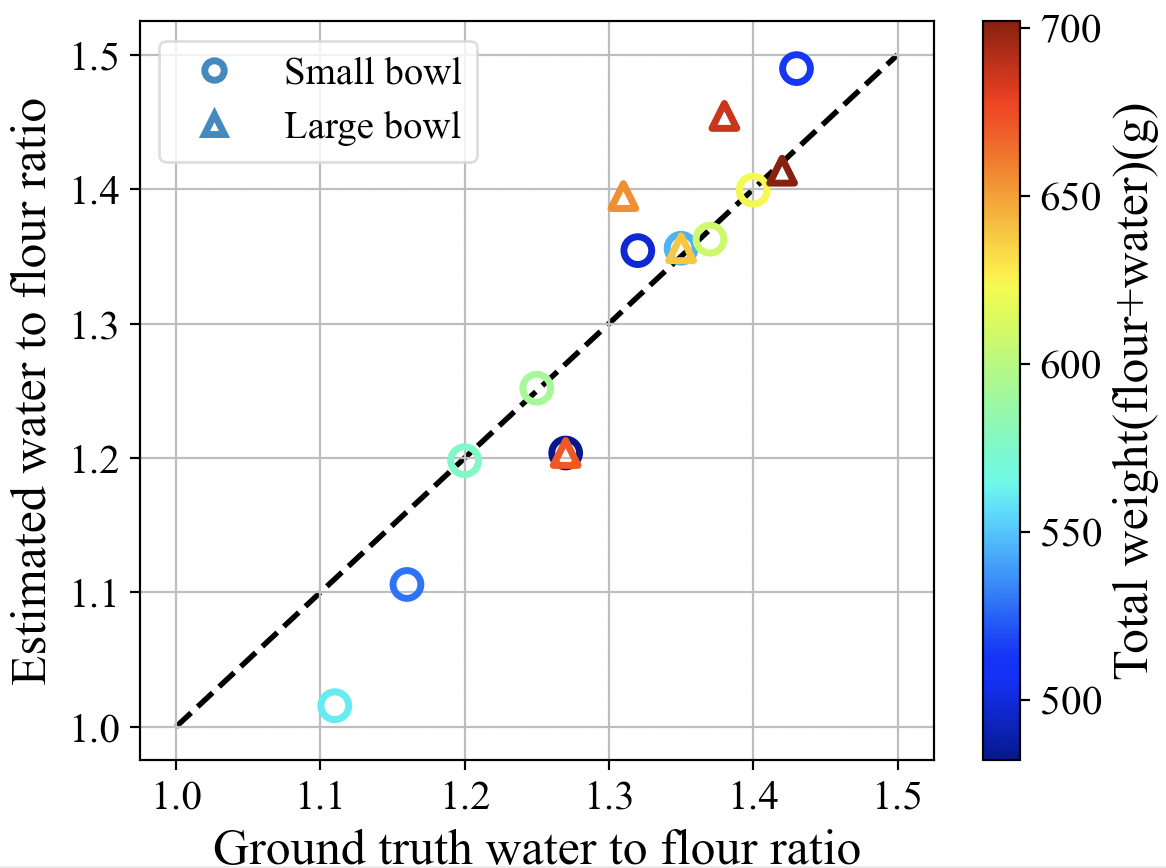}
\end{center}
\caption{Water-flour ratio estimation result of 15 batches of batter ($10$ in small bowl, $5$ in large bowl).}
\label{testset_info}
\vspace{0mm}
\end{figure}

\subsubsection{Uniformity Analysis}

Since uniformity has no official definition, we rely on assessments from humans with pancake cooking experiences to evaluate uniformity. We invited $5$ individuals to evaluate the final stirring outcomes of the $15$ batter batches of the test set. They can either observe it by eyes or utilize the whisk to feel the texture of the batter and assess its consistency. After observation, they were first asked to give a rating according to a scale ranging from $-3$ to $3$, where $-3$ indicates non-uniform batter and $3$ indicates over-uniform batter.
The mean uniformity score of $0.625$ with a variance of $1.1$ indicates that the stirring outcomes fall within the desired uniformity range. Results show that our stirring strategy and uniformity check also work well with a large bowl. 


\begin{table}[htbp]
    \centering
    \begin{tabular}{|c c c c c|}
        \hline
        \multicolumn{1}{|c}{} &
        \multicolumn{2}{c}{\textbf{Liquid level}} &
        \multicolumn{2}{c|}{\textbf{Water-flour ratio}} \\
        \hline
        \hline
       Bowl size & Small & Large & Small &Large\\
        \hline
        Average error & $1.37$ mm & $2.33$ mm & 0.033 & 0.048\\
        \hline
        Percentage error & $3.04\%$ & $3.88\%$ & $6.6\%$ & $9.6\%$\\
        \hline
    \end{tabular}
    \caption{Average and percentage errors in liquid level and water-flour ratio estimations for small and large bowls.}
    \label{liquid property estimation result table}
\end{table}

\subsubsection{Liquid level Estimation}
Table \ref{liquid property estimation result table} shows the error and percentage error of liquid level estimation on our test set. We have a total of 20 test trials whose liquid levels range from $5$ to $55$ mm. Liquid level error is as low as $3.04\%$. 



\subsubsection{Water-flour Ratio Estimation}
Table \ref{liquid property estimation result table} and Fig. \ref{testset_info} show the error and percentage error of water-flour ratio estimation on our test set. Although our testing error in the large bowl is slightly higher, it remains in an acceptable range, demonstrating the generalization potential of our method.


\subsection{Precise Pouring Control}

To make a pancake of the desired shape, we want to control the batter line strokes consistently and accurately when pouring.
In this section, we designed two tasks, \textbf{Line Stroke} and \textbf{Round Shapes},
to show that our method for line stroke drawing and round-shaped pancake control is precise. 
\subsubsection{Data Collection}
To form a shape, the robot pours the batter onto a 12-inch metal pizza plate, which we believe is the best available alternative to a traditional griddle. We use the RGB camera to capture and measure the results. We collected training and testing data of the two experiments and the test data did not replicate any scenarios from the training data, but the water-flour ratios were within the same range. The liquid level also varys for every trial of data we collected. 

\textbf{Line Strokes Task}:
We collect training data by controlling the robot arm moving speed and documenting the resulting line width. Our training dataset includes trials across five distinct water-flour ratios, ranging from $1.25$ to $1.45$. The resulting line widths of all experiments ranging from $0.8$ cm to $5.9$ cm. For the test data, we use five different batters, aiming to draw lines in four widths from $1$ cm to $4$ cm, which do not exist in training data. 

\textbf{Round Shapes Task}:
We collect training data by controlling the pouring time and documenting the resulting diameters. Our training dataset includes trials with the same range of water-flour ratios. The resulting pancake diameters ranging from $2$ cm to $22.5$ cm. For the test dataset, we use five different batters to pour four varying diameters ranging from $5$ cm to $20$ cm. 


\subsubsection{Model Implmentation}
We implement an MLP model with 1 hidden layer with 32 inputs and 64 outputs in PyTorch and train it using Adam optimizer for 1000 epochs and a learning rate of 0.06. The same network structure and setup are used for both experiments. Both MLP models are trained with MSE loss in a supervised manner. 

\begin{figure}[t]
\begin{center}
\vspace{2mm}
\includegraphics[scale=0.6]{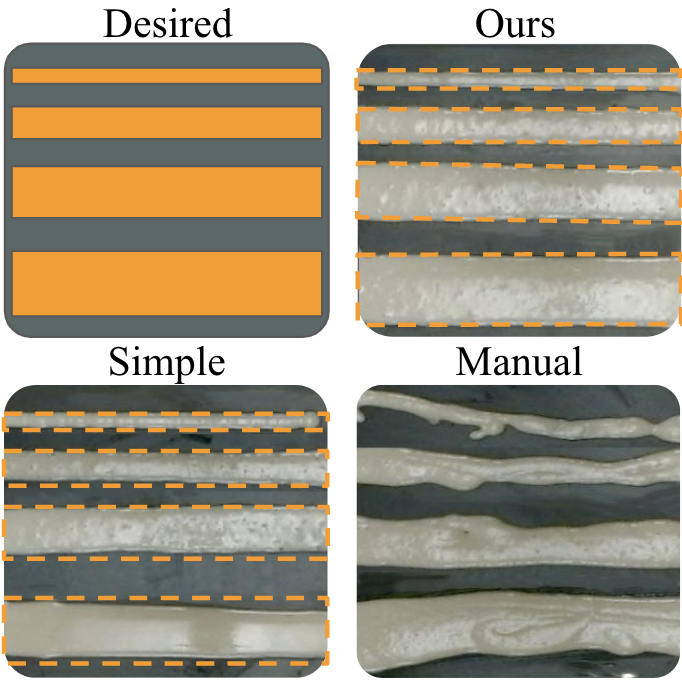}
\end{center}
\caption{A sample result of lines drawn using various control methods. The water-flour ratio for this result is 1.4}
\label{strokes_result}
\vspace{-5mm}
\end{figure}

\begin{figure}[t]
\begin{center}
\vspace{2mm}
\includegraphics[scale=0.28]{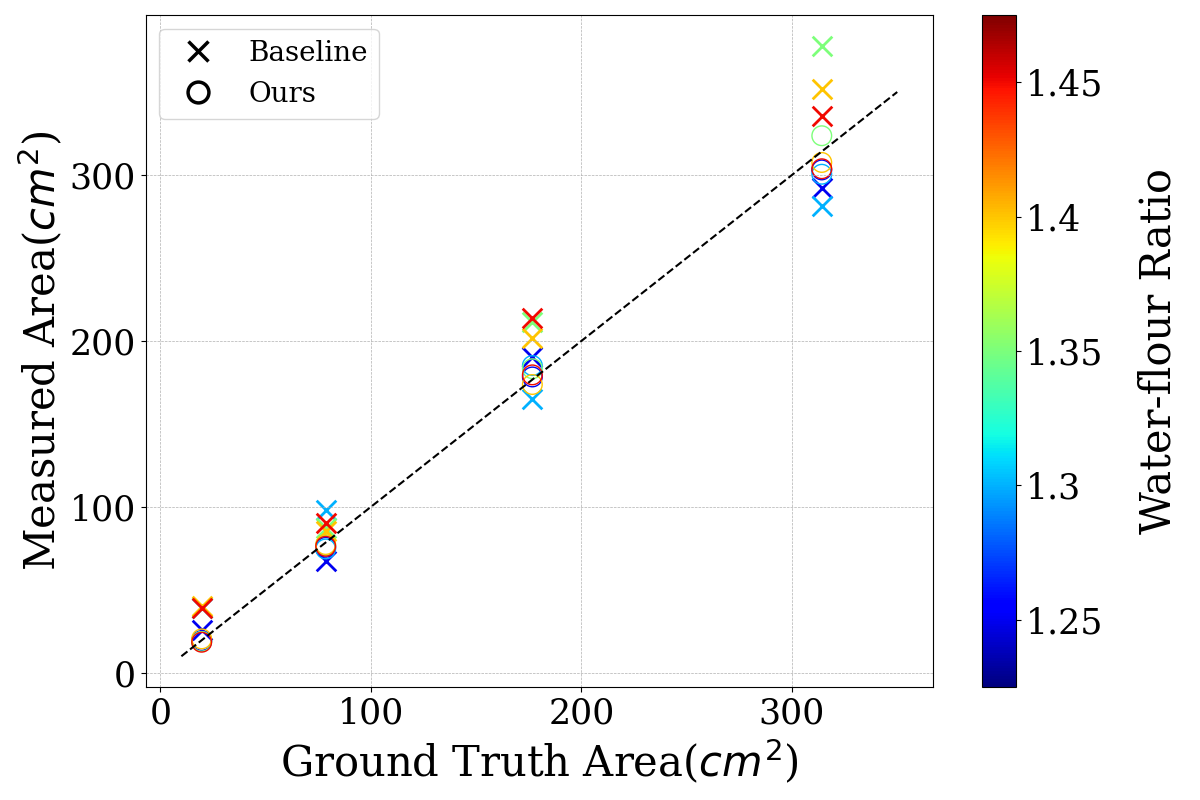}
\end{center}
\caption{Our round shape pancake pouring result using batter with different water-flour ratio. Our method pours pancakes with more accurate sizes than the baseline method.}
\label{round_graph}
\vspace{-5mm}
\end{figure}

\begin{table}[b]
\centering
\begin{tabular}{|l c c c|}
\hline
& \textbf{Ours} &\textbf{Simple} & \textbf{Manual} \\
\hline
\hline
Error ($mm$)  & \textbf{1.95} & 2.91  & 2.95   \\
\hline
Percentage Error & \textbf{9.5\%} & 12.7\% & 15.6\%\\
\hline
Variance ($mm$)  &  \textbf{1.93} & 2.03  & 4.71 \\

\hline
Percentage Variance & \textbf{9.6\%}& 9.9\% & 23.8\%\\

\hline
\end{tabular}
\caption{The table shows the results of the Line Strokes drawing task. Our method achieves both accuracy and consistency.}
\vspace{-5mm}
\label{table_line}
\end{table}

\begin{table}[hb]
\centering
\begin{tabular}{|l c c|}
\hline
& \textbf{Ours}& \textbf{Baseline} \\
\hline
\hline
Error ($cm^2$) &\textbf{5.25}  & 22.4   \\

\hline
Percentage Error& \textbf{3.88\%}  & 32.2\% \\

\hline
\end{tabular}
\caption{Comparison of round-shaped pancake area error and percentage error between Baseline and Our method. Ours performs much better than our baseline.}
\label{table_area}
\end{table}
\subsubsection{Baseline}
\hfill

\textbf{Line Strokes Task}:
The \textbf{Simple} method draws the line with the same arm moving speed regardless of the water-flour ratio of the batter. It assumes all batters have the same behavior. We also do not control the flow rate and assume the speed that is needed to draw a line for all water-flour ratios is the same. Moreover, we assume the speed of different widths is proportionally increasing. We found that the average speed needed to draw a line of $1$ cm wide is $0.2$ cm/s, then we increase this speed linearly for different line widths respectively.

The \textbf{Manual} method draws the line by a human expert holding the bowl. We asked the human expert to try to pour different water-flour ratio batter to draw the same width as our test set. We also give the human expert a reference width in the form of a piece of paper of the desired width.

\textbf{Round Shapes Task}:
The \textbf{baseline} method ignores the dynamic of the batter and only considers the target volume of pouring.
So we make the robot quickly reach the final angular position, which is determined by the volume we want to pour, and hold for $30$ seconds, where the vast majority of the batter is poured and only a few drops are left in some cases. 

\begin{figure}[t]
\begin{center}

\includegraphics[scale=0.18]{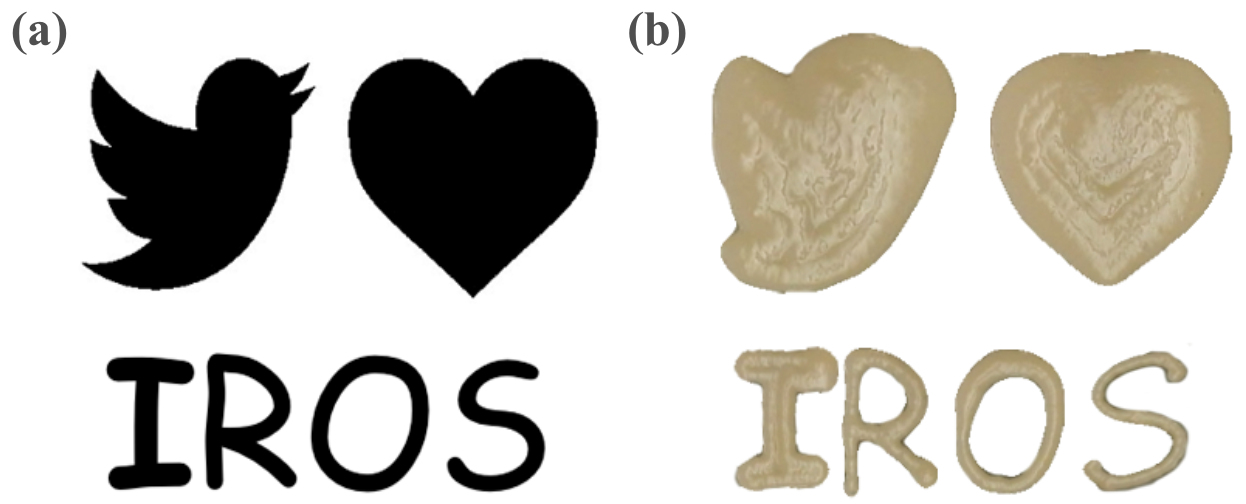}
\end{center}
\caption{We demonstrated our system can produce pancakes with arbitrary shapes. (a) shows the desired input images, and (b) shows the results achieved. }
\vspace{-5mm}
\label{demo}
\end{figure}
\subsubsection{Result}
We measured the width and the area difference of the poured batter from the desired shape. 
The result is shown in  Fig. \ref{strokes_result}, \ref{round_graph} and Table \ref{table_line}, \ref{table_area}. In summary, our method shows a significant improvement compared to the baseline where the batter properties are not considered.

\vspace{-2mm}

\subsection{Pancake Results of Various Shapes}
Finally, we tested the robot by creating different shapes of pancakes. 
We used our algorithm to decompose the image, stir the batter, and make the robot proceed pouring with the speed based on our model to maintain the desired width of the drawing stroke. 
Fig. \ref{demo} shows the results.

%% file: 06Conclusion.tex
\section{Conclusions And Future Works}
This paper presents a robotic system to perform effective pancake batter stirring and pouring of any arbitrary shape with force-torque and visual feedback. 
Our approach is composed of four stages, each dedicated to a specific aspect of pancake preparation or to gain more understanding of batter properties. The main goals of our systems include mixing the batter well, estimating properties such as water-flour ratio and liquid level accurately, and pouring the batter into the targeted shape precisely.
Our experiments and results show that our method can create a uniformly mixed batter, can control the poured line strokes with accuracy and consistency, and can poured into the desired shapes. However, our current system does not handle sharp turns in the image quite well, as shown in Fig. \ref{demo}. 
In the future, the work can be further improved if we can solve the problem with sharp edges in the desired images. One potential solution is with another method to form the shapes instead of pouring, to allow for more clear sharp turns and enhance the aesthetic appeal of the final pancakes.